# A Bottom-Up Approach for Automatic Pancreas Segmentation in Abdominal CT Scans

Amal Farag, Le Lu, Evrim Turkbey, Jiamin Liu and Ronald M. Summers

Imaging Biomarkers and CAD Laboratory, Department of Radiology and Imaging Sciences, National Institutes of Health Clinical Center, Bld. 10. Rm. 1C224D; Bethesda, MD

**Abstract.** Organ segmentation is a prerequisite for a computer-aided diagnosis (CAD) system to detect pathologies and perform quantitative analysis. For anatomically high-variability abdominal organs such as the pancreas, previous segmentation works report low accuracies when comparing to organs like the heart or liver. In this paper, a fully-automated bottom-up method is presented for pancreas segmentation, using abdominal computed tomography (CT) scans. The method is based on a hierarchical two-tiered information propagation by classifying image patches. It labels superpixels as pancreas or not via pooling patch-level confidences on 2D CT slices over-segmented by the Simple Linear Iterative Clustering approach. A supervised random forest (RF) classifier is trained on the patch level and a two-level cascade of RFs is applied at the superpixel level, coupled with multi-channel feature extraction, respectively. On six-fold cross-validation using 80 patient CT volumes, we achieved 68.8% Dice coefficient and 57.2% Jaccard Index, comparable to or slightly better than published state-of-the-art methods.

## 1 Introduction

Segmentation of the pancreas is an important step in the development of computer aided diagnosis (CAD) systems that can provide quantitative analysis for diabetic patients and a necessary input for subsequent methodologies for pancreatic cancer detection. The literature is rich for automatic segmentation of numerous organs in CT scans with high sensitivity (>90%), such as the liver, heart and kidneys. Yet, for segmentation of the pancreas, high accuracy in automatic segmentation remains a challenge. The pancreas shows high anatomical variations in shape, size and location that change from patient to patient. The amount of visceral fat tissue in the proximity can drastically vary the boundary contrast as well. All these factors make pancreas organ segmentation very challenging. Figure 1 shows different slices from three different patient cases and the ground-truth 3D segmented volumes of the pancreas to better visualize some of the variations and challenges mentioned.

Previous literature for pancreas segmentation in abdominal CT images are mostly **top-down** approaches that rely on atlas based approaches or statistical shape modeling or both [1-3]. In [1], Okada et. al perform multi-organ segmentation by combining inter-organ spatial interrelations with probabilistic atlases, which incorporates various a priori knowledge into the model, and a shape model to obtain results for seven organs. Shimizu et. al [2] utilize three-phase contrast-enhanced CT data which are first registered together for a particular patient and then registered to a reference patient by landmark-based deformable registration. A patient-specific probabilistic atlas guided segmentation is conducted, followed by an intensity-based classification and post-processing. The state-of-the-art result thus far is obtained by Wolz et. al [3]. A hierarchical weighted subject-specific atlas-based registration approach was implemented, with a Dice overlap using leave-one-out of 69.6% on 150 patients and 58.2% on a sub-population of 50 patients.

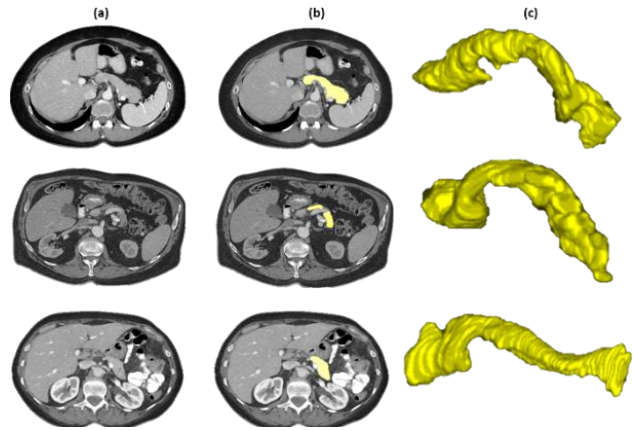

**Figure 1:** *Three color-coded masks in column (b) show the ground-truth pancreas segmentations for the slices in (a) of different patients. The corresponding 3D views are in (c).*

In this paper, a new **bottom-up** approach for pancreas segmentation is proposed with single phase CT patient data volumes. Our method is motivated to improve segmentation accuracy of highly deformable organs, such as the pancreas, by leveraging middle-level representation of image segments. Over segmentation of all 2D slices of a patient abdominal CT scan is first obtained as a semi-structured representation referred to as superpixels. The superpixel labeling maps are projected back to the 3D volumetric space. Random forest classifiers are trained once on the patch-level and in a two-level cascade fashion on the superpixel level, with multi-phase feature extraction processes, respectively.

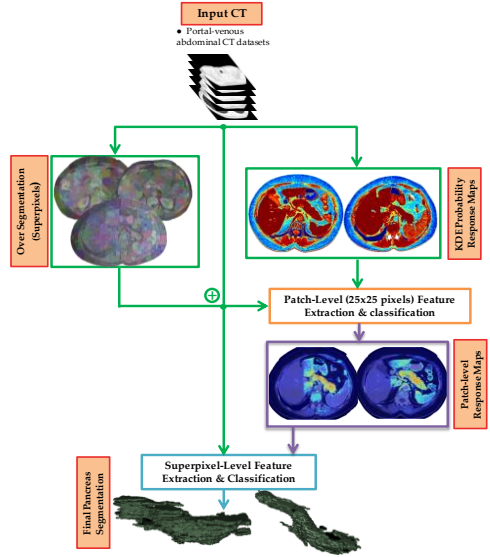

**Figure 2:** *Overall segmentation Framework.*

Leave-one-patient-out criterion is used as default in [1,3], for up to 150 patients. Here we would argue that leave-one-out based dense volume registration and label fusion process may be computationally impractical in clinical practice. More importantly, it does not scale up easily when large scale datasets are present. Thus, we employ 6-fold cross validation which exploits less data for training but more scans for testing. Classification models are compactly encoded by random forest classifiers through training, instead of label masks of (n-1) scans in leave-one-out of n patients. Our bottom-up approach is much more efficient than [1,3] in both memory and computation speed. In the literature a similar framework that utilizes superpixel methods and performs feature extraction and classification can be found for pathological region detection and segmentation within an organ [4]. MRI data is used and the overall feature extraction, classification and implementation details are significantly different from the proposed approach in this paper.

## 2 Methods

The overall bottom-up pancreas segmentation framework is illustrated in Figure 2. This section describes how to generate superpixels and each classification layer of our two-tiered approach in details.

### 2.1 Boundary-preserving over segmentation

There are two main broad categories of superpixel methods: gradient ascent and graph-based methods. Thorough examination and analysis of one gradient ascent and three graph-based superpixel algorithms are conducted, i.e., watershed [6], SLIC [5], efficient graph-based [7] and Entropy rate [8]. Evaluation is executed to find the most suitable set of parameters to obtain high boundary recalls (critical to the segmentation accuracy for the pancreas), in a range of distances of (1, 2, …, 6) pixels from the semantic pancreas ground-truth boundary annotation.

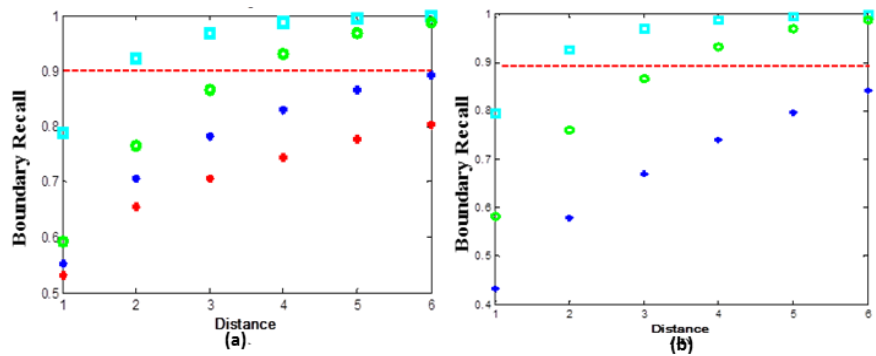

**Figure 3:** *Superpixels boundary recall results evaluated on a) 17 and b) 41 patient scans. The SLIC [5] results are represented in cyan, the watershed [6] in red, while the entropy rate [8] and efficient graph [7] based methods are depicted in green and blue, respectively. The red line represents the 90% marker.*

Quantitative and qualitative results can be found in Figures 3 and 4. Based on Figure 3, the SLIC approach provides the best boundary recall of >90%, under the distance of 2 pixels. The extension of superpixels to supervoxels is possible but we prefer 2D superpixel representation in this study, due to the potential boundary leakage problem of supervoxels deteriorating the pancreas segmentation more severely in multiple CT slices.

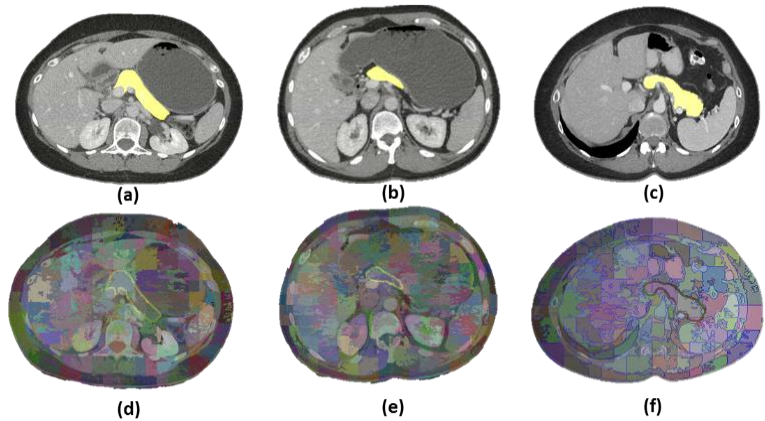

**Figure 4:** *Sample superpixels results from the SLIC method. First row are different slices from different patient scans with the ground-truth pancreas segmentation in yellow. The second row depicts the over segmentation results with the pancreas contours superimposed on the image.*

### 2.2 Patch-level feature extraction and labeling

This step consisted of two main components: feature extraction and classification. In the feature extraction stage, 14 different image features were computed and concatenated with additional dense D-SIFT [10], to capture fine-grained gradient or texture features, which results in 46 total features. Figure 6 shows the three sets of data information used for computations. The goal is to generate pancreas class-conditional response maps, as seen in Figure 6 (d, h).

The Scale Invariant Feature Transform [9] is a texture feature extractor and descriptor. In this paper, we adopt the Dense Scale Invariant Feature Transform (dSIFT) [10] which is based on SIFT [9] with several different extensions. Publically available VLFeat implementation is employed. In Figure 5, a sample image slice depicts the dSIFT process, where the descriptors are densely and uniformly extracted from image grids with inter-distances of 3 pixels. The green points on the image represent the patch center positions. Once the center positions are known, dSIFT is computed with bin size of 6 pixels and geometry of [2x2] bins, which results in 32 dimensional descriptors, for each image patch. For implementation and spatial bin configuration details, refer to [10]. The image patch size is chosen to be 25x25, a trade-off between the description power and computational efficiency. We empirically evaluated the size range from 15 to 35 pixels using small sub-sampled datasets for classification, as described later. Stable performance statistics are observed and we report quantitative experimental results using the default patch size of 25x25 pixels.

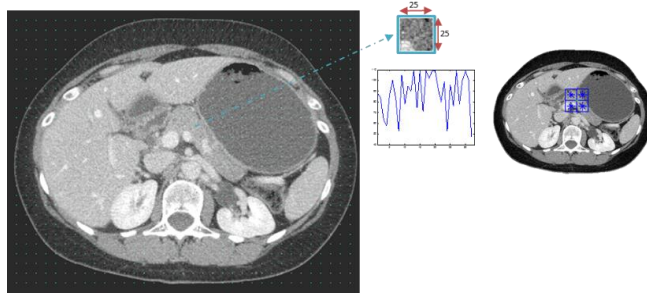

**Figure 5:** *Sample slice with center positions superimposed as green dots. The 25x25 image patch and corresponding D-SIFT descriptors are shown on the right-hand side.*

From the CT intensity modality, the mean, median and standard deviation (std) statistics over the full 25x25 pixel range per patch, $\mathcal{P}$, are extracted. The same intensity statistics within the intersected sub-region, $\mathcal{P}'$ of $\mathcal{P}$, and the underlying superpixel supporting mask (the superpixel where the patch center resides), obtained from Sec. 2.1, and the original patch are also extracted. The idea is that an image patch, $\mathcal{P}$, may be divided into more than one superpixel. The second set of statistics is calculated with respect to the most representative superpixel. In this manner, object boundary-preserving intensity features are obtained [12].

We also built similar features in the class-conditional probability density function (PDF) space. The ground-truth pancreas voxel intensities from 26 randomly selected patient CT scans were used as positive class samples and all remaining voxels were considered as negatives. These distributions were used to create kernel density estimators (KDE) that represent the CT intensity distributions of the positive $\{X^+\}$ and negative class $\{X^-\}$ of pancreas and non-pancreas voxels' CT image information. Let $X^+ = (h_1^+, h_2^+, \ldots, h_n^+)$ and $X^- = (h_1^-, h_2^-, \ldots, h_m^-)$ where $h_n^+$ and $h_m^-$ represent the intensity values for the positive and negative pixel samples for all 7 patient CT scans over the entire abdominal CT Hounsfield range. The kernel density estimators $f^+(X^+) = \frac{1}{n}\sum_{i=1}^{n} K(X^+ - X_i^+)$ and $f^-(X^-) = \frac{1}{m}\sum_{j=1}^{m} K(X^- - X_j^-)$ were computed where $K(\cdot)$ is assumed to be a Gaussian kernel. The normalized likelihood ratio was computed which becomes a probability value as a function of intensity in the range of $H = [0: 1: 4095]$. Thus, the probability of being considered pancreas is formulated as: $y^+ = \frac{f^+(X^+)}{f^+(X^+) + f^-(X^-)}$. Sample probability response maps are illustrated in Fig. 5 (c) and (g), where high probability regions are red in color and low probabilities in blue. In implementation, the above function can be converted as a pre-computed look-up table over $H = [0: 1: 4095]$, which allows very efficient $O(1)$ access time. The mean, median and std statistics are then computed from this normalized probability channel as well, with respect to $\mathcal{P}$ and $\mathcal{P}'$.

The overall 3D abdominal body region per patient can be reliably segmented using a standard table-removal procedure and all voxels outside the body are removed. We then compute the normalized relative x-axis and y-axis positions $\in [0, 1]$ at each of the image patch centroids, against the segmented body region. This provides the final two features extracted at the patch level for each axial slice in the patient volumes.

The total 46 patch level features were used to train the random forest (RF) classifier $C_1$. The classifier training was carried-out using six-fold cross validation. Figure 6 (d) and (h) show the computed response maps for the patch-level classification of two illustrative slices from different patients. The red color shows areas of high probability corresponding to the pancreas. From the response maps, the relative x and y positions as features are clearly important in separating positive and negative classes. The trained RF classifier is able to recognize the negative class patches residing in the background, such as liver, vertebrae and muscle using spatial location cues. For example, note the implicit vertical and horizontal decision boundary lines in Fig. 6 (d, h). Comparing Fig. 6 (d) and (h) versus (c) and (g) respectively, it demonstrates the superior descriptive and discriminative power of the feature descriptor on image patches ($\mathcal{P}$ and $\mathcal{P}'$) than single pixel intensities. Organs with similar CT values are significantly depressed in the patch-level response maps.

In summary, SIFT and its variations, e.g., D-SIFT have shown to be informative, especially through spatial polling or packing [13]. Our defined 14 features also capture a wide range of visual information and pixel-level correlations per image patch. Both good classification recall and specificity have been obtained in cross-validation using Random Forest implementation of 50 trees (i.e., the treebagger( ) function in Matlab). In future work, we plan to exploit the deep convolutional neural network based approach for dense patch labeling [16], without pre-defined features.

**2.3 Superpixel-level feature extraction and classification**

In this stage, the 2D superpixel supporting maps (recording the spatial partitioning using SLIC), the original CT image slices and the probability response maps from the patch classification are used for feature extraction on a superpixel level. Treating the collection of CT voxels and the per-voxel/patch response values (from $C_1$) within any superpixel as two empirical unordered distributions, higher 1~4 order statistics such as mean, std, skewness, kurtosis [14] and 7 percentiles (20%, 30%, 40%, …, 90%) were computed. This results in 24 features for each instance that are used to train a cascade of two random forest classifiers. A cascade of random forests was employed here, due to the highly unbalanced quantities between foreground (pancreas) superpixels and background (the rest of CT volume) superpixels, which is general for rare event detection [15]. The superpixel labels are inferred from the overlapping ratio $r$ of the superpixel label map and the ground-truth object level pancreas mask. If $r \geq 0.5$, the superpixel is labeled as positive; if $r \leq 0.2$, negative. For a small portion of superpixels with $0.5 > r > 0.2$, they are ambiguous to assign labels and thus not used for training. A two-level cascaded random forest classification hierarchy was found to be sufficient and implemented to obtain $C_2$ and $C_3$. Figure 7 shows the receiver operating curves (ROC) for 6-fold cross validation. From the AUC values, $C_3$ is harder to train since it employs the hard negatives as negative samples but classified positively by $C_2$. RF with 50~200 trees are evaluated, with similar empirical performances.

The binary 3D pancreas volumetric mask is simply obtained by stacking the binary superpixel classification/labeling outcomes from 2D axial slices. No further post-processing is employed since the major focus of this paper is to investigate the performance effects of using superpixels as a middle-level representation on organ seg-mentation. Post-processing could improve the segmentation accuracy, such as 3D morphological operators, connected component analysis, surface-based level-set or graph-cut based MRF/CRF optimization. This is left for future work.

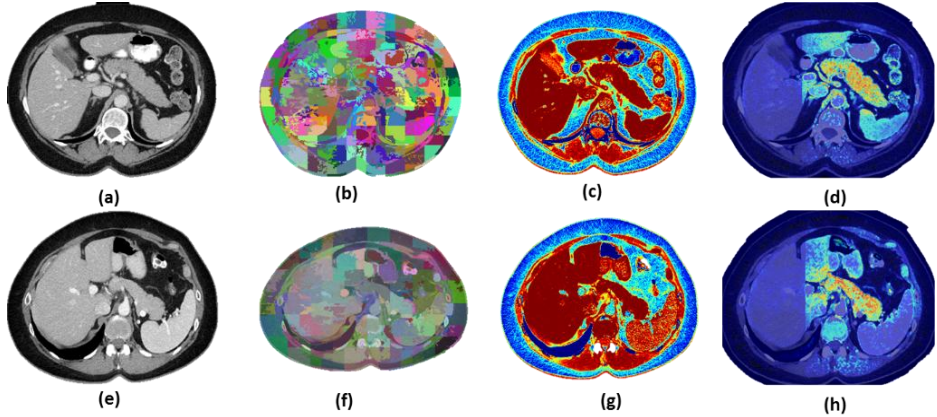

**Figure 6:** *Two sample slices from different patients are shown in (a) and (e). The corresponding superpixels segmentation (b,f), probability response maps (c, g) and patch-level probability response maps (d, h) are shown. In (c,g) and (d,h), red represents highest probabilities. In (d,h) the purple color represents areas where probabilities are so small and can be deemed insignificant areas of interest.*

## 3  Experimental Results and Validation

**Data & Metrics:** We use 80 annotated abdominal CT patient scans to assess the accuracy and robustness measures for pancreas segmentation using the proposed method. 17 of the subjects are from a kidney donor transplant list of healthy patients that have abdominal CT scans prior to kidney extraction. The remaining 63 patients are selected by a radiologist from the Picture Archiving and Communications System (PACS) that had neither pancreatic cancer lesions nor major abdominal pathologies. The 80 datasets are acquired from different CT scanners in the portal-venous phase (~70s after intravenous contrast injection) with slice thickness ranging from 1.5-2.5 mm with tube voltage 120 kV. Manual ground-truth segmentations of the pancreas for all 80 cases are provided by a medical student and verified/modified by a radiologist. Several similarity metrics were calculated to validate the accuracy and robustness of our method. The Dice similarity index is used to interpret the overlap between two sample sets, $SI = 2\frac{|A \cap B|}{|A|+|B|}$ where $A$ and B refer to the algorithm output or manual ground-truth 3D pancreas segmentation. The Jaccard index (JI) is another statistic used to compute similarities between the segmentation result against the reference standard, $JI = \frac{|A \cap B|}{|A \cup B|}$. The volumetric recall (i.e. sensitivity) and precision values are also reported.

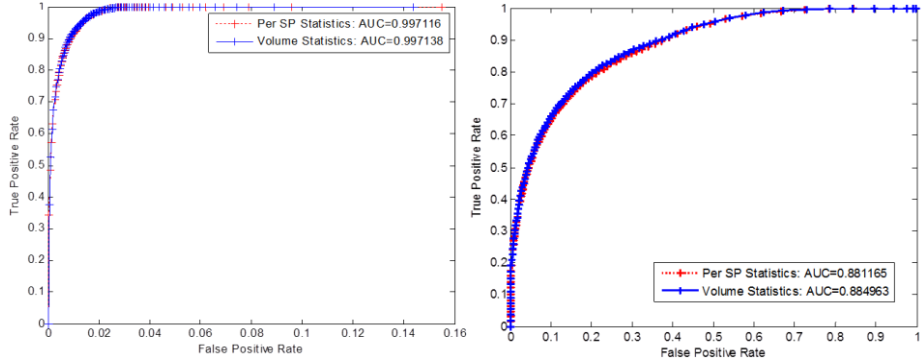

**Figure 7:** *ROC curves to analyze the superpixel classification results, in a two layer cascade of RF classifiers: (a) the first layer classifier, $C_2$ and (b) the second layer classifier, $C_3$. Red plots are using each superpixel as a count to calculate sensitivity and specificity. In blue plots, superpixels are weighted by their size (e.g., numbers of pixels) for calculation.*

Our results are evaluated on different numbers of patients (see Table 1): 41, 60 and 80, respectively, which demonstrates how performance changes with the additions of more patients data. Steady improvements of ~4% in the Dice coefficient and ~5% for the Jaccard index are observed, from 41 to 60, and 60 to 80. Our best segmentation result (Dice of 68.8% at 80 patients in 6-fold cross validation) is closely approaching the highest accuracy level of [3] (Dice of 69.6% at 150 patients in leave-one-out), with less than 1% difference. At 41 patients, our result is 2.2% better than [3] with 50 patients (Dice coefficients of 60.4% versus 58.2%). Leave-one-out validation translates to performing computationally demanding 3D dense non-rigid registration and label fusion of numerous volumes to the target case, in order to obtain the segmentation of one patient. However, larger standard deviations in performance measurements are observed for our method. In summary, our bottom-up segmentation approach is a more computationally efficient (2~3 minutes versus 30 minutes) method that demonstrates comparable results against the state-of-the-art [3], using 6-fold cross validation instead of leave-one-out [3].

**Table 1:** *Comparison with state-of-the-art segmentation methods. Average Dice overlap (Similarity Index, SI), Jaccard Index (JI) and Recall/Precision*

| Reference | N Patients | Dice (Similarity Index) | Jaccard Index | Precision | Recall |
|---|---|---|---|---|---|
| *Wolz et. al* | 50 | 58.2%±20.0 [0 81.2] | 43.5%±17.8 [0 68.6] | | |
| *Wolz et. al* | 150 | 69.6%±16.7 [6.9 90.9] | 55.5%±17.1 [3.6 83.3] | 67.9%±18.2 [6.0 91.8] | 74.1%±17.1 [8.0 93.4] |
| *Proposed* | 41 | 60.4%±22.3 [2.0 96.4] | 46.7%±22.8 [0 93.0] | 55.6%±29.8 [1.2 100] | 80.8%±21.2 [4.8 99.8] |
| *Proposed* | 60 | 64.9%±22.6 [0 94.2] | 51.7%±22.6 [0 89.1] | 70.3%±29.0 [0 100] | 69.1%±25.7 [0 98.9] |
| *Proposed* | 80 | 68.8%±25.6 [0 96.6] | 57.2%±25.4 [0 93.5] | 71.5%±30.0 [0 100] | 72.5%±27.2 [0 100] |

**Discussion:** Our protocol is arguably harder than the leave-one-out criterion in [3,1] since less patient datasets are exploited in training and more separate patient scans for testing. In fact, [3] does demonstrate a notable performance drop from using 149 patients in training versus 49 patients, i.e., mean Dice coefficients decreased from

69.6% to 58.2%. This indicates that the multi-atlas fusion approaches [3,1] may actually achieve lower segmentation accuracies than reported, if under six-fold cross validation.

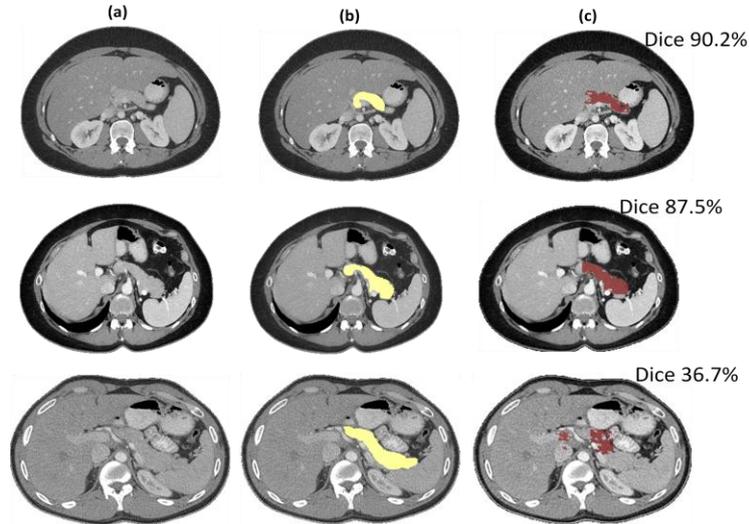

**Figure 8:** *Segmentation results with computed good, fair and poor Dice coefficients for the pancreas. Sample original slices for three patients are shown in (a) and the corresponding groundtruth manual segmentations in (b) are in yellow. Final segmentations are shown in red in (c) with Dice coefficients for the volume above each slice.*

Figure 8 shows samples of final pancreas segmentation results for three different patients, where good result refers to computed Dice coefficient above 90% (15 patients), fair result as $50\% \leq Dice \leq 90\%$ (49 patients) and poor for $Dice < 50\%$ (16 patients).

## 4 Conclusions

In this paper a hierarchical two-tiered method was proposed based on a bottom-up information propagation from image patches to segments. The SLIC superpixel generation algorithm [5] provided the best overall pancreas organ-level boundary recall by partitioning each 2D CT axial slice into over-segmentation label maps of all patients. Their final binary labeling masks can be straightforwardly stacked and projected back into the 3D CT scan space, to form the pancreas segmentation mask. Random forest classifier and cascade of RF classifiers were trained at the image patch- and superpixel-level respectively, via extracting multi-channel features. Based on a six-fold cross validation of the 80 CT datasets, our results are comparable and slightly better than the state-of-the art work [3,1]. A Dice coefficient of 68.8% and Jaccard index of 57.2% was obtained, versus 69.6% and 55.5% in [3] and JI of 46.6% [1]. Further analysis of the approach will be conducted on employing additional datasets and exploiting superpixel based 2D/3D random field models, i.e., taking superpixels as nodes to form intra- and inter-slice graph connections, for more structurally sophisticated post-processing.

*Acknowledgements:* A.A.F is an Imaging Sciences Training Program Fellow. This research was supported by the Intramural Research Programs of the National Institutes of Health Clinical Center and National Institute of Biomedical Imaging and Bioengineering.